\def\year{2018}\relax
\documentclass[letterpaper]{article} 
\usepackage{rotating}
\usepackage{aaai18}  
\usepackage{times}  
\usepackage{helvet}  
\usepackage{courier}  
\usepackage{url,natbib}  
\usepackage{graphicx}  
\usepackage{booktabs} 
\usepackage{multirow}
\usepackage{pgfplots}
\usepackage{tikz}
\usepackage{tkz-tab}
\usepackage{subcaption}
\usepackage{amsmath}
\usepackage{amsfonts}

\begin{document}
%
\title{Unity in Diversity: Learning Distributed Heterogeneous Sentence Representation for Extractive Summarization}

\author{Abhishek Kumar Singh, Manish Gupta\thanks{The author is also a Principal Applied Researcher at Microsoft.} and Vasudeva Varma\\
 IIIT Hyderabad, India\\
 abhishek.singh@research.iiit.ac.in, manish.gupta@iiit.ac.in, vv@iiit.ac.in
 }

\maketitle
\begin{abstract}
Automated multi-document extractive text summarization is a widely studied research problem in the field of natural language understanding. Such extractive mechanisms compute in some form the worthiness of a sentence to be included into the summary. While the conventional approaches rely on human crafted document-independent features to generate a summary, we develop a data-driven novel summary system called HNet, which exploits the various semantic and compositional aspects latent in a sentence to capture document independent features. The network learns sentence representation in a way that, salient sentences are closer in the vector space than non-salient sentences. This semantic and compositional feature vector is then concatenated with the document-dependent features for sentence ranking. Experiments on the DUC benchmark datasets (DUC-2001, DUC-2002 and DUC-2004) indicate that our model shows significant performance gain of around 1.5-2 points in terms of ROUGE score compared with the state-of-the-art baselines. 
\end{abstract}

\section{Introduction}
The rapid growth of online news over the web has generated an epochal change in the way we retrieve, analyze and consume data. The readers now have access to a huge amount of information on the web. For a human, understanding large documents and assimilating crucial information out of it is often a laborious and time-consuming task. Motivation to make a concise representation of huge text while retaining the core meaning of the original text has led to the development of various automated summarization systems. These systems provide users filtered, high-quality concise content to work at unprecedented scale and speed. Summarization methods are mainly classified into two categories: \textit{extractive} and \textit{abstractive}. Extractive methods aim to select salient phrases, sentences or elements from the text while abstractive techniques focus on generating summaries from scratch without the constraint of reusing phrases from the original text.

The majority of literature on text summarization is dedicated to extractive summarization approach.
Previous methods can be predominantly categorized as (1) greedy approaches (e.g.~\citep{carbonell1998use}), (2) graph based approaches (e.g.~\citep{erkan2004lexrank}) and (3) constraint optimization based approaches (e.g.~\citep{mcdonald2007study}). These approaches rely mainly on a set of features which were manually crafted. Recently, few efforts have been made towards data-driven learning approaches for extractive summarization using neural networks.~\citet{kaageback2014extractive} used recursive autoencoders to summarize documents, achieving good performance on the Opinosis~\citep{ganesan2010opinosis} dataset.~\citet{cao2015learning} used convolution neural networks for addressing the problem of learning summary prior representation for multi-document extractive summarization.~\citet{cheng2016neural} introduced attention based neural encoder-decoder model for extractive single document summarization trained on a large corpus of news articles collected from the Daily Mail. Their work focuses on sentence-level as well as the word-level extractive summarization of individual documents using encoder-decoder architecture.~\citet{DBLP:conf/cikm/Singh0V17} proposed a combination of memory network and convolutional BLSTM (Bidirectional Long Short Term Memory) network to learn better unified document representation which jointly captures n-gram features, sentential information and the notion of the summary worthiness of sentences leading to better summary generation. 

Most successful multi-document summarization systems use extractive methods. Sentence extraction is a crucial step in such a system. The idea is to find a representative subset of sentences, which contains the information of the entire set. Thus, sentence ranking is imperative in finding such an informative subset, which sets our focus to sentence-level summarization. The performance of the summarization system using sentence ranking approach is profoundly determined by the feature engineering, irrespective of the ranking models~\citep{osborne2002using,conroy2004left,galley2006skip,li2007multi}. Features are broadly classified as: (a) document-dependent features (e.g., position, term frequency), and (b) document-independent features (e.g., length, stop-word ratio, word polarity). Document independent features often reveal the aspect that a sentence can be considered summary worthy irrespective of which document it is present in. Consider the following example.
\begin{enumerate}
\item Six killed, eight wounded in a shooting at Quebec City.
\item It was the shooting that killed six people and injured eight people at a Quebec City mosque.
\end{enumerate}
While the former sentence conveys prominent information in concrete terms, the latter is a more verbose way of portrayal with similar meaning. In the case of multi-document summary, the former sentence is the best candidate, as it is a concise representation keeping important information intact. This intuition was called as summary prior nature by ~\citet{cao2015learning}, and can be captured by learning better document independent features.


We aim to learn a better sentence representation that incorporates both document dependent features as well as document independent features to capture the notion of saliency of a sentence. Since the sentence representation comprises of two different kinds of features, we call it a heterogeneous representation. Contrary to the orthodox method of painstakingly engineering document independent features, we propose a model with a Convolutional Sentence Tree Indexer (CSTI), a novel data-driven neural network for capturing semantic and compositional aspects in a sentence. CSTI slides over the input sequence to produce higher-level representation by compressing all the input information into a single representation vector of the root node in the constructed binary tree. We present details in Section~\ref{propmodel}. Final sentence representation obtained by concatenating the transformed document dependent features and the features obtained from CSTI (document independent features) is used under a regression framework for sentence ranking.

Deep neural networks perform better in the case of huge training data. However, non-availability of large multi-document summarization corpus makes learning challenging for deep networks and often results procured are not of high quality. To overcome this issue, we use transfer learning approach where we first train the network on single document summarization corpus~\citep{cheng2016neural} and then fine-tune the network with the multi-document datasets. We summarize our key contributions below.

\begin{enumerate}
\item We propose CSTI, a novel method to encode semantic and compositional features latent in a sentence which can be combined with document dependent features to learn a better heterogeneous sentence representation for capturing the notion of summary worthiness of a sentence.
\item Further, we propose a novel Siamese CSTI (Siam-CSTI) model for effectively identifying redundant sentences during the sentence selection process.  
\item We use transfer learning method to overcome the problem of lack of data for multi-document summarization.
\item We experimentally show that our method outperforms the basic systems and several competitive baselines. Our model achieves significant performance gain on the DUC 2001, 2002 and 2004 multi-document summarization datasets.
\end{enumerate}

\section{Related Work}
Extractive document summarization has been traditionally connected to the task of sentence ranking. Sentence ranking models by~\citet{osborne2002using,conroy2004left,galley2006skip,li2007multi} are dependent on the human-crafted features.~\citet{shen2007document} modeled extractive document summarization as a sequence classification problem using Conditional Random Fields. Our approach is different from theirs as we use a data-driven approach to automatically acquire document-independent features for representing sentences without the need of manually crafted document independent features.~\citet{hong2014improving} built a summarization system using advanced document-independent features which can be seen as an attempt to capture better sentence representation. These features are often hand-crafted and fail to capture various semantic aspects. Summarization system CTSUM~\citep{wan2014ctsum} attempts to rank sentences using certainty score. However, certainty score alone is not enough to reveal all possible latent semantic aspects.~\citet{DBLP:conf/coling/RenWCMZ16} develop a redundancy aware sentence regression framework for multi-document extractive summarization. They model importance and redundancy simultaneously by evaluating the relative importance of a sentence given a set of selected sentences. Along with single sentence features they incorporate additional features derived from the sentence relations. They manually crafted \textit{sentence importance features} and \textit{sentence relation features} while we use deep neural network for getting automatic document-independent features.

Recursive Neural Networks are known to model compositionality in natural language over trees. The tree structure is predefined by a syntactic parser~\citep{socher2013recursive} and each non-leaf tree node is associated with a node composition function.~\citet{socher2013recursive} also proposed Tensor networks as composition function for sentence level sentiment analysis tasks. Recently,~\citet{zhu2015long} introduced S-LSTM which extends LSTM units to compose tree nodes in a recursive fashion. Neural Tree Indexer (NTI), an extension of S-LSTM was proposed for natural language inference and QA task~\citep{DBLP:conf/eacl/YuM17}. In our work we introduce a CSTI, an enhanced version of NTI adapted for summarization task. Unlike NTI, our model uses (a) CNNs that can slide over inputs to produce higher-level representations, and (b) BLSTM as the primary composition function. 



\section{Proposed Model}
\label{propmodel}
\begin{center}
\begin{figure}
\includegraphics[width=3.5in]{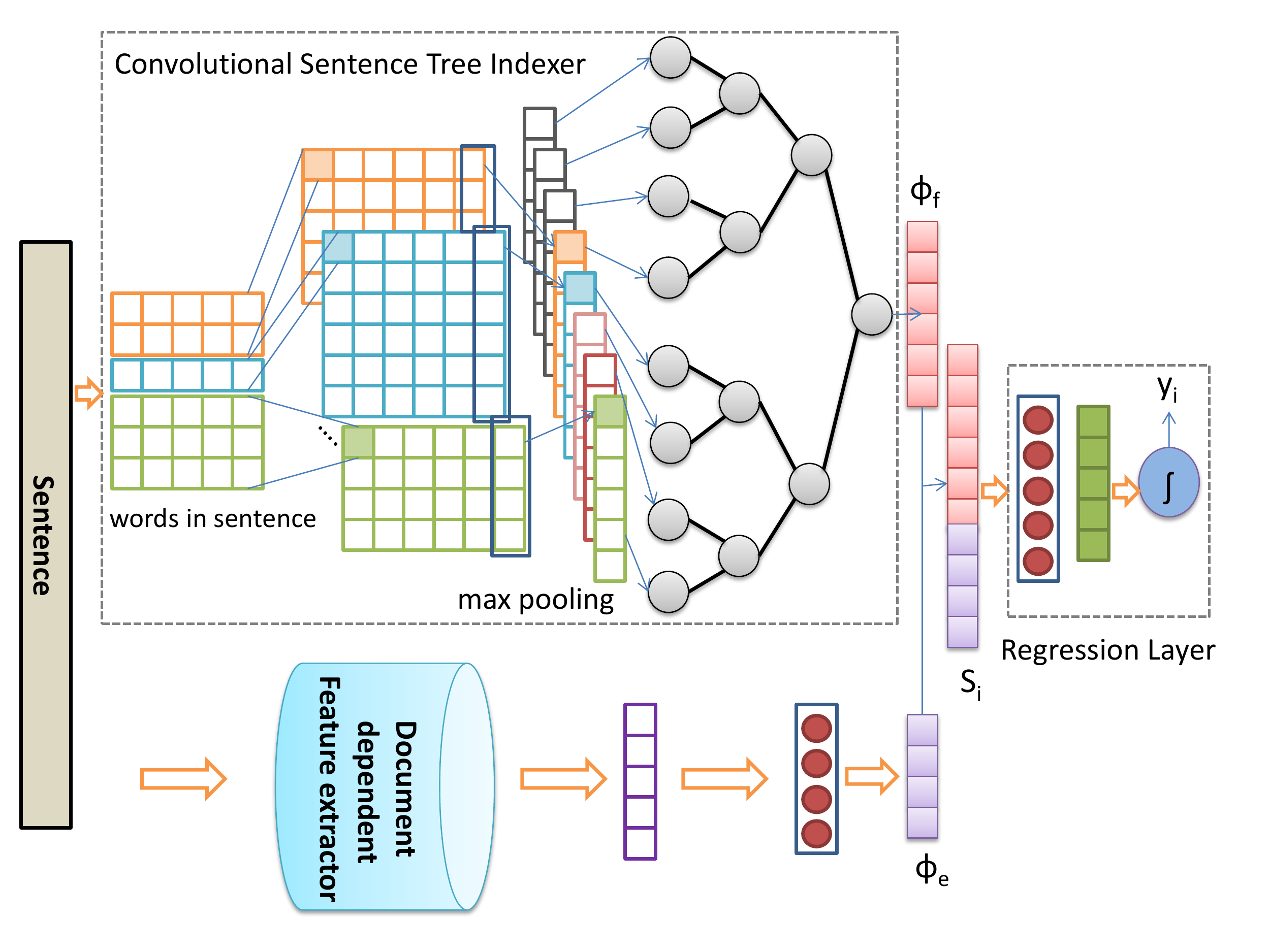}
\caption{The System Architecture of HNet. After max pool operation padding vectors (represented in black color) are added to form a full binary tree.}
\label{systemarch}
\end{figure}
\end{center}

Our architecture intends to learn a better representation of a sentence with consideration of both document-dependent and document-independent features in order to measure the worthiness of a sentence in the summary. The proposed system architecture is illustrated in Figure~\ref{systemarch}.  
The principal components of our model architecture are as follows.
\begin{enumerate}
\item \textit{CSTI}: captures local (word n-grams and phrase level), global (sequential and compositional dependencies between phrases) information and the notion of saliency of a sentence. Details in Section~\ref{subsec:csti}.
\item \textit{Extractor}: extracts document dependent features from the given sentence. Details in Section~\ref{extractor}.
\item \textit{Regression Layer}: predicts sentence scores and thus, helps in the sentence ranking process.
\end{enumerate}
\noindent
CSTI provides an embedding which incorporates document-independent features. Final unified sentence embedding is obtained by concatenating embedding from CSTI and document-dependent features, which is then forwarded through the regression layer to obtain saliency score of a sentence. Since the model makes use of the heterogeneous representation of the sentence, we name our model as Heterogeneous Net (HNet). In this section, we first describe the CSTI and then present details of the extractor and the regression layer.

\subsection{Convolutional Sentence Tree Indexer (CSTI)}
\label{subsec:csti}
We focus on learning a hierarchical sentence representation that not only incorporates phrase level features and global sentence level information but it should also include the notion of saliency of a sentence. The hierarchical nature of our model reflects the fact that sentences are generated from words, phrases and often have some sequential and compositional dependencies among these units. Therefore, we use an architecture to obtain a representation with minimum information loss such that the global information gets discovered and the local information remains preserved.

CSTI comprises of: (a) \textit{Convolutional Encoder}: We use a Convolution Neural Network (CNN) with multiple filters to automatically capture set of phrase (n-grams) based features followed by a max-over-time pooling operation to obtain a set of feature vectors. We do this because phrases with different lengths can exhibit the same characteristics of summary prior nature. (b) \textit{Bidirectional Long Short Term Memory Tree Indexer (BLSTM Tree Indexer)} to obtain a comprehensive set of document-independent features incorporating semantic and compositional aspects in a sentence. We use BLSTM Tree Indexer because: (a) it models conditional and compositional power of sequential RNNs and syntactic tree based recursive neural nets, and (b) it is a robust syntactic parsing-independent tree structure model and does not require a parse tree structure. 
\subsubsection{Convolutional Encoder}
For first level sentence encoding, we choose convolution neural network for the  following reasons: (1) it is easily trainable without long-term dependencies, (2) it handles sentences of variable length inherently and is able to learn compressed representation of n-grams effectively, (3) previous research has shown that it can be successfully used for sentence-level classification tasks such as sentiment analysis~\citep{kim2014convolutional}.



Conventional convolution neural network uses convolution operation over various word embeddings which is then followed by a max pooling operation. Suppose, $d$ dimensional word embedding of the $i^{th}$ word in the sentence is $w_i$, and let $w_{i:i+n}$ denote the concatenation of word embeddings $w_i, ..., w_{i+n}$. Then, convolution operation over a window of $c$ words using a filter of $\theta_t^{c} \in \mathbb{R}^{m\times cd} $ yields new features with $m$ dimensions. Convolution operation is written as follows.
\begin{equation}
	\label{eq1}
    f_i^{c} = tanh(\theta_t^{c} \times w_{i:i+c-1} + b)
\end{equation}
Here $b$ is the bias term. We obtain a feature map $F^c$ by applying filter $\theta_t^{c}$ over all possible windows of $c$ words in the sentence of length $N$.
\begin{equation}
	\label{eq2}
    F^{c} = [f_1^{c}, f_2^{c}, ..., f_{N-c+1}^c] 
\end{equation}

Our intention is to capture the most prominent features in the feature map. Hence, we used max-over-time pooling operation~\citep{collobert2011natural} to acquire final features for a filter of fixed window size. To exploit several latent features from phrase based information, we used multiple filters of different window widths. Let $\theta_t^1, \theta_t^2, ..., \theta_t^k$ be $k$ filters for window sizes from 1 to $k$ then we have $k$ feature maps $F^1, F^2, ..., F^k$. Applying max-over-time pooling operation helps to get most salient features. They seem to capture the phrase-level information nicely. The first level features $\phi_1$ obtained from convolution network can be denoted as follows.
\begin{equation}
	\label{eq3}
    \phi_1 = \{max\{F^1\}, max\{F^2\}, ..., max\{F^k\}\}
\end{equation}

We use an enhanced convolution network which is different from the one used for sentence classification task~\citep{kim2014convolutional} or for learning the prior summary task~\citep{cao2015learning}.~\citet{kim2014convolutional} reserves all representation generated by filters to a fully connected layer which ignores relations among phrases with different lengths.~\citet{cao2015learning} tried to capture this relation by performing two-stage max-over-time pooling operation. Unlike these models, our model captures the relation among different length phrases by passing the representations generated after max-over-time pooling operation to the BLSTM Tree Indexer network. Representation thus obtained also incorporates the latent temporal and compositional dependencies among variable length phrases.

\subsubsection{BLSTM Tree Indexer (BTI)}
Sequential LSTMs are known to learn syntactic structure (conditional transition) from natural language. However their generalization to unseen text is relatively poor in comparison with models that exploit syntactic tree structure~\citep{DBLP:conf/nips/BowmanMP15}. BLSTM Tree Indexer leverages the sequential power of LSTMs and the compositional power of recursive models, without the need of a parse tree. The model constructs a binary tree
by processing the input sequences with its node function in a bottom-up fashion. It compresses all the input information into a single representation vector of the root node. This representation seems to capture both semantic and compositional aspects in the sentence.

The output of the convolutional encoder is padded with padding vectors to form a full binary tree and fed as input to the BLSTM Tree Indexer. The input set consists of a sequence of vectors ($\phi_1$). BTI can be a full n-ary tree structure. To reduce computational complexity, we have implemented binary tree form of BTI in  our study. It has two types of transformation functions: (a) a non-leaf node composition function $f^{node}(h^1 ,...,h^q)$ and (b) a leaf node transformation function $f^{leaf}(\phi^{j}_{1})$, where $\phi^{j}_1$ is $j^{th}$ feature vector from set $\phi_1$. $f^{node}(h^1, ..., h^q)$ is a composition function of the representation of its child nodes $h^1, ..., h^q$, where $q$ is the total number of child nodes of this non-leaf node. $f^{leaf}(\phi^{j}_{1})$ is some non-linear transformation of the input vector $\phi^{j}_1$.

As we use the binary tree form of BTI, a non-leaf node can only take two direct child nodes, i.e., $q = 2$. Hence, the function $f^{node}(h^l, h^r)$ learns a composition over its left child node $h^l$ and right child node $h^r$. The node and the leaf node functions are actually parameterized neural networks. 

We present our approach for the two types of transformation functions in the following.

\noindent
\textbf{\textit{Leaf Node Transformation:}}
We use a MLP (Multi-Layer Perceptron) with $ReLU$ function (for non-linear transformation) for the leaf node function $f^{leaf}$ as follows.
\begin{equation}
h_j = ReLU(MLP(\phi^{j}_1 ; \theta))
\end{equation}

\noindent where $\phi^{j}_1$ is input sequence fed to the multi-layer perceptron, $\theta$ is the learning parameter and $h_j$ is the vector representation for the leaf node.

\noindent
\textbf{\textit{Non-Leaf Node Composition:}}
A Bidirectional LSTM (BLSTM) is used as the composition function $f^{node}(h^l, h^r)$ to get the representation of the parent node. BLSTM processes the input both in the forward order as well in the reverse order, allowing to combine future and past information in every time step. It comprises of two LSTM layers processing the input separately to produce $\overrightarrow{h}$, $\overrightarrow{c}$, the hidden and cell states of an LSTM processing the input in the forward order, and $\overleftarrow{h}$ and $\overleftarrow{c}$, the hidden and the cell states of an LSTM processing the input in reverse order. Both, $\overrightarrow{h}$ and $\overleftarrow{h}$, are then combined to produce output sequence of the BLSTM layer. Let $h^{l}_{t}$, $h^{r}_{t}$, $c^{l}_{t}$ and $c^{r}_{t}$ be the vector representations and cell states for left and right children. A BLSTM computes a parent node representation $h^{p}_{t+1}$ and a node cell state $c^{p}_{t+1}$ as follows.
\newline
Forward order:
\begin{equation}
	\label{eq4}
 \overrightarrow{i_{t+1}} = \sigma(W_{1}\overrightarrow{h^{l}_{t}}+W_{2}\overrightarrow{h^{r}_{t}}+W_{3}\overrightarrow{c^{l}_{t}})+W_{4}\overrightarrow{c^{r}_{t}}
\end{equation}
\begin{equation}
	\label{eq5}
 \overrightarrow{f^{l}_{t+1}} = \sigma(W_{5}\overrightarrow{h^{l}_{t}}+W_{6}\overrightarrow{h^{r}_{t}}+W_{7}\overrightarrow{c^{l}_{t}})+W_{8}\overrightarrow{c^{r}_{t}}
\end{equation}
\begin{equation}
	\label{eq6}
 \overrightarrow{f^{r}_{t+1}} = \sigma(W_{9}\overrightarrow{h^{l}_{t}}+W_{10}\overrightarrow{h^{r}_{t}}+W_{11}\overrightarrow{c^{l}_{t}})+W_{12}\overrightarrow{c^{r}_{t}}
\end{equation}
\begin{equation}
	\label{eq7}
 \overrightarrow{c^{p}_{t+1}} = \overrightarrow{f^{l}_{t+1}}\odot \overrightarrow{c^{l}_{t}}+\overrightarrow{f^{r}_{t+1}}\odot \overrightarrow{c^{r}_{t}}+\\ \overrightarrow{i_{t+1}}\odot \tanh(W_{13}\overrightarrow{h^{l}_{t}}+W_{14}\overrightarrow{h^{r}_{t}})
\end{equation}
\begin{equation}
\label{eq8}
\overrightarrow{o_{t+1}} = \sigma(W_{15}\overrightarrow{h^{l}_{t}}+W_{16}\overrightarrow{h^{r}_{t}}+W_{17}\overrightarrow{c^{p}_{t+1}})
\end{equation}
\begin{equation}
\label{eq9}
\overrightarrow{h^{p}_{t+1}} = \overrightarrow{o_{t+1}}\odot \tanh(\overrightarrow{c^{p}_{t+1}})
\end{equation}
\noindent
Similarly, in the reverse order we obtain $\overleftarrow{h^{p}_{t+1}}$ and $\overleftarrow{c^{p}_{t+1}}$. Finally, we combine them to obtain the vectors $c^{p}_{t+1}$ and $h^{p}_{t+1}$ as follows.
\begin{equation}
\label{eq10}
\nonumber c^{p}_{t+1} = mean(\overrightarrow{c^{p}_{t+1}}, \overleftarrow{c^{p}_{t+1}}),\ \ \  h^{p}_{t+1} = mean(\overrightarrow{h^{p}_{t+1}}, \overleftarrow{h^{p}_{t+1}})
\end{equation}
\noindent
where $W_1$, ..., $W_{17}\in \mathbb{R}^{k\times k}$ ($k=N-c+1$) are trainable weights. For
brevity we eliminated the bias terms. $\sigma$ and $\odot$ denote the element-wise sigmoid function and the element-wise vector multiplication respectively. Each non-leaf node computes its representation by composing its children representation using the above set of equations. This representation is passed towards the root in a bottom-up fashion to construct the tree representation. The vector representation of the root $h^{root}$ (also referred as $\phi_f$) incorporates semantic and compositional aspects latent in a sentence.

\subsection{Extractor}
\label{extractor}
Besides just using the document independent features, we also intend to use document dependent features in learning better sentence representation for saliency estimation of sentences.
Our feature set includes the following: (1) The position of the sentence, (2) The averaged cluster frequency values of words in the sentence, (3) The average term frequency values of the words in the sentence, (4) The average word IDF values in the sentence, divided by sentence length, and (5) The maximal word IDF score in the sentence.


We choose these features for the following reasons: (1) They tend to impart some contextual knowledge. (2) They are often simple to calculate and have been extensively used in previous research~\citep{cao2015ranking,cao2015learning}.

\subsection{Regression Layer}
We follow traditional supervised learning approach for sentence ranking~\citep{carbonell1998use,li2007multi}. The regression layer at the end of the architecture aims to assign scores to a sentence. Sentences are ranked based on the score predicted by the regression layer. Since our approach focuses on learning a better sentence representation embracing both document-independent and document-dependent features, we concatenate the document-independent features obtained from the CSTI net with the transformed extracted document-dependent features (using a dense layer). Let $\phi_e$ be the transformed extracted document-dependent features and let $S_i$ denote the heterogeneous sentence embedding of the $i^{th}$ sentence. Thus, $S_i = [\phi_f, \phi_e]$.

\noindent The sentence worthiness is scored by ROUGE-2~\citep{lin2003automatic} (without stop words) and our model tries to estimate this score. Given sentence $i$, the final sentence representation $S_i$ is used in the regression layer to score saliency as $y_i = \sigma(W^T \times S_i)$ where $W$ are the regression weights and $\sigma$ is the softmax function. The softmax function gives a nice distribution over the range $[0, 1]$ which makes it suitable to imitate ROUGE score.  

This score is used to rank sentences. Higher the score, higher is the chance of the sentence to be included in the generated summary.


\begin{center}
\begin{figure}[t]
\includegraphics[height=2in, width=3.5in]{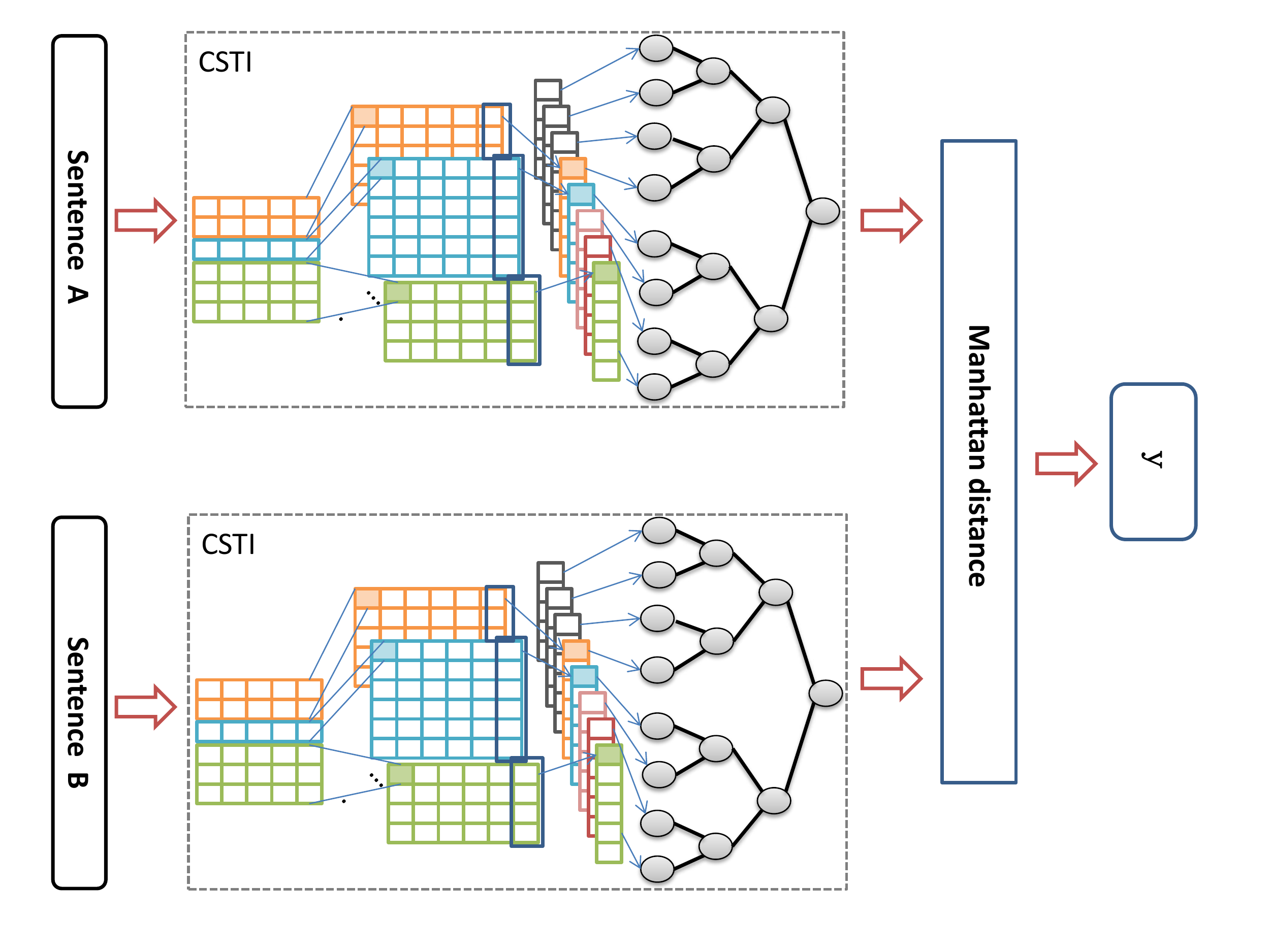}
\caption{Siamese CSTI Architecture}
\label{siam-csti}
\end{figure}
\end{center}

\subsection{Removing Redundant Sentences}
A good summary should be informative with non-redundant content. We generate the final summary by choosing top ranked sentences taking into account the redundancy among the selected sentences. The sentences are sorted in descending order of saliency scores. To identify whether the next candidate sentence is redundant, we compare it with all the sentences in the summary generated so far. We introduce a Siamese CSTI (Siam-CSTI) network for identifying redundant sentences. Figure~\ref{siam-csti} shows the Siam-CSTI architecture. The base network consists of the CSTI net. Weight parameters are tied for the base network. Two CSTI nets feed their output to a distance metric layer. We experiment with cosine, Euclidean and Manhattan distances and empirically find that the Manhattan distance seems to perform better in our case.
 
Siam-CSTI network is trained for sentence similarity task on the SICK data~\citep{DBLP:conf/semeval/MarelliBBBMZ14}. We present dataset details in Section~\ref{subsec:datasets}. Formally, we consider a supervised learning setting where each training example consists of a pair of sequences
$(x^{a}_{1}, ..., x^{a}_{T_a})$, $(x^{b}_{1}, ..., x^{b}_{T_b})$
of fixed-size vectors (each $x^{a}_i$, $x^{b}_j$ $\in \mathbb{R}^{d}$ is d-dimensional word vector) along with a single label $y$ for the pair.
The sequences may be of different lengths $T_a \neq T_b$ and the sequence lengths can vary from example to example. The similarity function $g$ is based on the Manhattan distance metric as follows.
\begin{equation}
\label{siam-eqn}
g(h^{a}_{T_a}, h^{b}_{T_b}) = \exp(-||h^{a}_{T_a}-h^{b}_{T_b}||_1) \in [0, 1]
\end{equation}
\noindent where $h^{a}_{T_a}$, $h^{b}_{T_b}$ are the learned representations of the sequences $x^{a}_{T_a}$, $x^{b}_{T_b}$ respectively such that $h^{a}_{T_a}$ and $h^{b}_{T_b}$ are closer in the vector space if $x^{a}_{T_a}$ and $x^{b}_{T_b}$ are similar otherwise they reside far apart. Mean Squared Error (MSE) is used as loss function (after rescaling the training set relatedness labels to lie in $[0, 1]$). The Siam-CSTI model trained on paired examples seems to learn a highly structured space of sentence representations by exploiting the sequential and recursive power of CSTI that captures rich semantics. Similar sentences ($y=1$) are considered as redundant sentences and non-similar sentences ($y=0$) are considered as non-redundant sentences. The final summary is generated by iteratively picking up a sentence from the set of previously sorted sentences and adding it to current summary if the picked sentence is non-redundant.

\section{Experimental Setup}
We experiment with our CSTI and Siam-CSTI  based summarization model (HNet) for the task of multi-document summarization. In this section, we present our experimental setup for assessing the performance of our system. We discuss the corpora used for training and evaluation and provide implementation details of our approach.

\begin{table}[t]
\begin{center}
\scriptsize
\begin{tabular}
{|p{5.8cm}|p{0.3cm}p{0.3cm}p{0.3cm}|}
\hline
\bf Ranking by Dependency Tree-LSTM & \bf Tree & \bf M & \bf S  \\\hline\hline
\bf a woman is slicing potatoes& & & \\
$\bullet$ a woman is cutting potatoes
&4.82   &4.87& 4.91\\
$\bullet$ potatoes are being sliced by a woman&4.70   &4.38& 4.68\\
$\bullet$ tofu is being sliced by a woman&4.39   &3.51& 3.62\\ \hline
\bf a boy is waving at some young runners from
the ocean & & &\\
$\bullet$ a group of men is playing with a ball on the beach&3.79 &3.13 & 2.68\\
$\bullet$ a young boy wearing a red swimsuit is jumping
out of a blue kiddies pool&3.37 &3.48 & 3.29\\
$\bullet$ the man is tossing a kid into the swimming pool
that is near the ocean&3.19&  2.26& 1.87\\

\hline
\end{tabular}
\end{center}
\caption{\label{siam-sim}Most similar sentences (from 1000-sentence subsample) in the SICK test data according to the Tree-LSTM.
Tree/M/S denote relatedness (with the sentence preceding
each group) predicted by the Tree-LSTM/MaLSTM/Siam-CSTI.}
\end{table}








\subsection{Datasets}
\label{subsec:datasets}
Initial training of our model is done on the Daily Mail corpus, used for the task of single document summarization by~\cite{cheng2016neural}. 
Overall, we have 193986 training documents, 12147 validation documents and 10350 test documents in the corpus. For the purpose of training, we created a sentence and its ROUGE-2 score pairs from this corpus. Sentences which are part of the summary get high ROUGE scores than non-summary sentences. We experiment on DUC 2001-2004 datasets which are used for generic multi-document summarization task.
These documents are from newswires which are grouped into several thematic clusters. The full DUC data set can be availed by request at {\tt http://duc.nist.gov/data.html}. 
The DUC 2001, 2002 and 2004 datasets consist of 11295, 15878 and 13070 sentences respectively. The SICK dataset which contains 9927 sentence pairs with a 5,000/4,927 training/test split~\citep{DBLP:conf/semeval/MarelliBBBMZ14} was used for training the Siam-CSTI net. Each pair has a relatedness label $\in [1, 5]$ corresponding to the average relatedness judged by 10 different individuals.


\subsection{Implementation Details}
We fine tuned our model on DUC datasets after initial training on Daily Mail corpus. DUC 2003 data is used as development set and we perform a 3-fold cross-validation on DUC 2001, 2002 and 2004 datasets with two years of data as training set and one year of data as the test set. The word vectors were initialized with 250-dimensional pre-trained embeddings~\citep{mikolov2013distributed}. The embeddings for ``out of vocabulary'' words were set to zero vector. The size of the hidden units of BLSTM was set to 150. After tuning on the validation set, we fix the dimension $m$ of the latent features from convolutional encoder as 125 and window size $k = 5$ for HNet system. 
We use Adam~\citep{kingma2014adam} as the optimizer with mini batches of size 35. Learning rates are set to \{0.009, 0.0009\}. For our network, we use regularization dropout of \{0.2, 0.5\}.


\subsection{Baseline Methods}
\label{baseline}
In this section of the paper, we describe several summarization baseline systems that we choose to compare against our system. These baselines include best peer systems (PeerT, Peer26, and Peer65) which participated in DUC data evaluations, state-of-the-art summarization results on DUC 2001, 2002 and 2004 corpus respectively. We select the systems that performed best on DUC 2001, 2002, 2004 datasets, which are: (1) R2N2~\citep{cao2015ranking}, (2) ClusterCMRW~\citep{wan2008multi}, (3) REGSUM~\citep{hong2014improving}, (4) PriorSum~\citep{cao2015learning}, and (5) RASR~\citep{DBLP:conf/coling/RenWCMZ16}.
The R2N2 system uses a recursive neural network to rank sentences by automatically learning to weigh hand-crafted features. ClusterCMRW system leverages the cluster-level information and incorporates this information into a graph-based ranking algorithm. REGSUM follows a word regression approach for doing better estimation of word importance which leads to better extractive summaries.  PriorSum captures summary prior nature by exploiting phrase based information. RASR uses regression framework that simultaneously learns the model importance and redundancy information by calculating the relative gain of a sentence with respect to given set of selected sentences. Further, we use LexRank~\citep{erkan2004lexrank} as a baseline to compare performance level of regression approaches. We also compare with StandardCNN and Reg\_Manual. StandardCNN consists of just conventional CNNs with fixed window size for learning sentence representation. Reg\_Manual is used as a baseline system to explore and understand the effects of learned sentence representation prior to the summary. It adopts human-compiled document-independent features: (a) NUMBER (if a number exists), (b) NENTITY (if named entities exist), and (c) STOPRATIO (the ratio of stopwords). It combines these features with document dependent features and tunes the feature weights through LIBLINEAR\footnote{\url{ http://www.csie.ntu.edu.tw/~cjlin/liblinear}} support vector regression.

\begin{table*}[t]
\begin{center}
\small
\begin{tabular}{|lll|lll|lll|}
\hline 
\multicolumn{3}{|c|}{2001}& \multicolumn{3}{|c|}{2002}& \multicolumn{3}{|c|}{2004}\\
\hline
\bf System & \bf ROUGE-1 & \bf ROUGE-2&\bf System & \bf ROUGE-1 & \bf ROUGE-2&\bf System & \bf ROUGE-1 & \bf ROUGE-2 \\ 
 PeerT &33.03&7.86&Peer26 &35.15&7.64&Peer65 &37.88&9.18\\
 R2N2 &35.88&7.64& ClusterCMRW &38.55&8.65& REGSUM &38.57&9.75\\
 LexRank &33.43&6.09& LexRank &35.29&7.54& LexRank &37.87&8.88\\
 Reg\_Manual &35.95&7.86& Reg\_Manual &35.81&8.32& Reg\_Manual &38.24&9.74\\
 StandardCNN &35.19&7.63& StandardCNN &35.73&8.69& StandardCNN &37.9&9.93\\
 PriorSum &35.98&7.89& PriorSum &36.63&8.97& PriorSum &38.91&10.07\\
 RASR &36.31&8.49& RASR &37.8&9.61& RASR &36.6&10.57\\
 HNet-B &36.82&8.64& HNet-B &38.79&9.43& HNet-B &39.27&10.85\\
 HNet-B(T)&37.69&9.12& HNet-B(T) &39.52&9.69& HNet-B(T) &39.9&11.08\\
 HNet&37.21&8.96& HNet &39.17&9.61& HNet &39.54&10.94\\
 HNet$^{-}$&34.51&7.88& HNet$^{-}$ &35.86&8.24& HNet$^{-}$ &35.66&9.37\\
 HNet(T)& \bf 38.18 & \bf 9.43 & HNet(T) & \bf 39.94 & \bf 9.92   & HNet(T) & \bf 40.34& \bf 11.29  \\
\hline
\end{tabular}
\end{center}
\caption{\label{res}Comparison Results (\%) on DUC Datasets}
\end{table*}

%
%

\section{Results and Analysis}
\label{reseval}
In this section, we compare the performance of our system against various summarization baselines using ROUGE-1 (unigram match) and ROUGE-2 (bigram match) measures. We also attempt to analyze our system trained with different approaches with intuition and empirical evidence presented in the form of tables and graphs. Lastly, we conclude this section by presenting examples of sentences selected for summaries by the proposed system.

We carried out extensive experiments with diverse settings in order to evaluate our system. In doing so, we created several variations of HNet model which are: (1) HNet-B: uses Convolutional BLSTM as sentence encoder instead of CSTI. (2)  HNet-B(T): refers to HNet-B model which is trained with transfer learning approach, i.e., the model was first trained on Daily Mail dataset~\citep{cheng2016neural} and was then fine-tuned on multi-document DUC datasets. (3) HNet: refers to our proposed model with CSTI as sentence encoder for sentence ranking and Siam-CSTI as redundancy identifier for sentence selection task. (4) HNet(T): refers to HNet model which was first trained on Daily Mail dataset~\citep{cheng2016neural} and was then fine-tuned on multi-document DUC datasets. (5) HNet$^{-}$: refers to the HNet model when the embedding from the extractor ($\phi_e$) is made zero.

It is evident from the results presented in Table~\ref{res} that our basic systems HNet-B and HNet-B (T) significantly outperform (T-test with p-value=0.05) state-of-the-art summarization systems R2N2, Cluster-CMRW, REGSUM, PriorSum, and RASR. 
This is encouraging because despite having not so complex deep network architecture the HNet-B system is able to learn efficient document-dependent semantic features. It also outperforms the Reg\_Manual baseline which uses human-compiled features for obtaining the document-independent features and the graph-based summarization system LexRank. From the results shown in Table~\ref{res} it is clear that HNet-B outperforms the StandardCNN baseline. This is due to the fact that the additional BLSTM network used in HNet-B helps in learning temporal (sequential) dependencies among variable length phrases exploiting past as well as future context. Finally, our proposed model (HNet/HNet (T)) significantly outperforms our basic systems HNet-B/HNet-B (T) which supports the fact that HNet is equipped with suitable deep network architecture for procuring latent semantic features (document-independent features) from a sentence. In the following, we analyze different aspects of the proposed system.

\noindent\textbf{Contribution of Document Independent Features}
\noindent
To explore the contribution of the learned document independent features towards the saliency estimation of a sentence prior to the summary, we follow a simple approach. For each sentence, we ignore document dependent features by setting the $\phi_e$ vector to \textbf{0}, and then applying the regression transform to calculate the saliency score. We refer to this model as HNet$^{-}$. This setting helps us in analyzing the intuitive features latent in our heterogeneous representation of the sentence without consideration of the contextual features. After comparing results of HNet$^{-}$ and HNet in Table~\ref{res}, we observe a difference of around 3--4 points and 1--2 points in terms of ROUGE-1 and ROUGE-2 scores respectively. The drop in points has resulted due to the absence of document dependent features. Therefore, we can conclude that document independent features have a major contribution towards saliency estimation of a sentence. This experiment also supports the need of document dependent features as incorporating them results in significant increase in ROUGE scores as provided in Table~\ref{res}.

\noindent\textbf{Significance of BTI in CSTI (HNet Model)}
\noindent After performing rigorous experiments, we observe that the use of BTI as part of CSTI significantly enhances the performance of the HNet system. This fact is evident when we compare HNet performance against StandardCNN and PriorSum as they use only CNN for obtaining semantic representation of a sentence. The performance improvement is better reflected in the case of HNet(T) system because of increase in the training data. Adding BLSTM Tree Indexer increases the number of parameters to be learned in the network. The more the training data the better the robustness of the system.
HNet also outperforms (T-test with p-value=0.04) HNet-B. This is due to the fact that BTI constructs a full binary tree by processing the input sequence with its node functions in a bottom-up fashion. It compresses all the input information into a single representation vector of the root. This representation seems to capture the sequential and recursive dependencies among various units (words/phrases) of the sentence.

\noindent\textbf{Significance of Siam-CSTI in Sentence Selection} From Table~\ref{siam-sim} it is evident that Siam-CSTI performs better (T-test with p-value=0.02) than similar state-of-the-art architectures: TreeLSTM~\citep{DBLP:conf/acl/TaiSM15} and MaLSTM~\citep{DBLP:conf/aaai/MuellerT16} for sentence similarity task. We also experimented with basic TF-IDF cosine similarity and empirically found the superior performance of Siam-CSTI. The network seems to exploit the sequential and recursive aspects of the sentences to learn a rich set of semantics that help in identifying similar sentences.   

\noindent\textbf{Contribution of Transfer Learning Method}
The fact that increase in training data results in better performance as the system becomes more robust motivated us to pre-train the HNet-B and HNet systems on Daily Mail dataset first and then fine-tune the systems to multi-document summarization setting. We refer to these systems as HNet-B(T) and HNet(T). Table~\ref{res} shows the improvement in results for these systems in terms of ROUGE-1 and ROUGE-2 scores on DUC benchmark datasets. HNet(T) is the best performing system amongst the HNet variants.
\noindent\textbf{Examples of Sentences Selected by HNet(T)}: In Table~\ref{vis}, we provide examples of some high scored sentences and low scored sentences selected by our HNet(T) system. From Table~\ref{vis}, we observe that the learned representation high-scores the sentences that consist of more facts (named entities, numbers etc.) and low-scores the sentences that contain more stop-words and/or are informal and so often fail to provide rich facts.






\setlength{\tabcolsep}{2pt}
\begin{table}[h]
\scriptsize
\begin{center}
\begin{tabular}{|l|p{8cm}|}
\hline
\multirow{3}{*}{\rotatebox{90}{High  scored\ \ \ \ }}
& $\bullet$ The largest tanker spill in history resulted from the July 19, 1979, collision off Tobago of the supertankers Atlantic Empress and Aegean Captain, in which 300,000 tons  more than 80 million gallons of oil was lost.\\
& $\bullet$ If the approximate 200,000 illegal aliens were not counted, the county would loose an estimated $\$56$ million a year in federal revenue and lose representatives in Congress.\\
\hline
\multirow{3}{*}{\rotatebox{90}{Low scored\ \ \ \ }}
& $\bullet$ His coach and physician had also testified at the inquiry. \\
& $\bullet$ The House had twice rejected efforts to exclude aliens.\\
& $\bullet$ However, that oil burned as well as spilled.\\
& $\bullet$ The new growth will attract a larger variety of birds and other animal life to the area.\\

\hline
\end{tabular}
\end{center}
\caption{\label{vis}Example Sentences Selected by HNet(T)}
\end{table}

\section{Conclusions}
We proposed a novel deep neural network to learn sentence representations combining both document-dependent and document-independent aspects. The architecture consists of a CSTI which acts as a sentence encoder, an extractor module which extracts document dependent features, a Siam-CSTI net which identifies redundant sentences, and a regression layer which performs sentence saliency scoring. The proposed system discovers various inherent semantic and compositional aspects as part of document-independent features. We also showed that the use of transfer learning approach helps in overcoming the learning issues faced by the network due to the shortage of training data for multi-document summarization. Experimental results on DUC 2001, 2002, and 2004 datasets confirmed that our system outperforms several state-of-the-art baselines.


\bibliographystyle{aaai} \bibliography{aaai}

\end{document}